\documentclass[journal]{IEEEtran}
\usepackage{amsmath,amsfonts}
\usepackage{algorithmic}
\usepackage{algorithm}
\usepackage{array}
\usepackage[caption=false,font=normalsize,labelfont=sf,textfont=sf]{subfig}
\usepackage{textcomp}
\usepackage{stfloats}
\usepackage{url}
\usepackage{verbatim}
\usepackage{graphicx}
\usepackage{cite}
\usepackage{booktabs}
\usepackage{multirow}
\usepackage{amssymb}
\usepackage[colorlinks=true, linkcolor=blue, citecolor=black, urlcolor=magenta]{hyperref}
\begin{document}

\title{TP-JSCC: Efficient Token Pruning and Joint Source-Channel Coding for LVLMs-based 3D-VQA in Cloud-Edge Collaboration}

\author{Zhengxin Li, 
        Chongzhen Tian, 
        Hui Yuan,~\IEEEmembership{Senior Member,~IEEE}, 
        Hao Liu, 
        and Sam Kwong,~\IEEEmembership{Fellow,~IEEE}

\thanks{This work was supported in part by the National Natural Science Foundation of China under Grant 62571303. \textit{(Zhengxin Li and Chongzhen Tian contributed equally to this work.) (Corresponding author: Hui Yuan.)}}

\thanks{Zhengxin Li and Chongzhen Tian are with the School of Control Science and Engineering, Shandong University, Jinan 250061, China, and also with the Key Laboratory of Machine Intelligence and System Control, Ministry of Education, Jinan 250061, China (e-mail: 202520831@mail.sdu.edu.cn; cztian@mail.sdu.edu.cn).}

\thanks{Hui Yuan is with the School of Control Science and Engineering, Shandong University, Jinan 250061, China, and also with the Shandong Key Laboratory of Intelligent Communication and Sensing-Computing Integration, Jinan 250100, China (e-mail: huiyuan@sdu.edu.cn).}

\thanks{Hao Liu is with the School of Computer and Control Engineering, Yantai University, Yantai, 264005, China (e-mail: liuhaoxb@gmail.com).}


\thanks{Sam Kwong is with the School of Data Science, Lingnan University, Hong Kong (e-mail: samkwong@ln.edu.hk).}
}

\markboth{Journal of \LaTeX\ Class Files,~Vol.~14, No.~8, August~2021}%
{Shell \MakeLowercase{\textit{et al.}}: A Sample Article Using IEEEtran.cls for IEEE Journals}


\maketitle

\begin{abstract}
Deploying large vision-language models (LVLMs) on resource-constrained embodied agents for 3D visual question answering necessitates cloud-edge collaboration. However, offloading massive visual data to the cloud over bandwidth-limited and noisy channels introduces severe latency and transmission distortion. To address this, we propose TP-JSCC, an efficient token pruning and joint source-channel coding method for LVLMs-based 3D-VQA in Cloud-Edge Collaboration. First, we develop a Local and Global Semantics-based Hybrid Token Pruning (LGS-HTP) module to significantly reduce the visual token count. By decoupling foreground-background tokens and jointly selecting local and global semantics, LGS-HTP effectively reduces redundancy while preserving reasoning-critical tokens. Second, we introduce a lightweight Token Joint Source-Channel Coding (T-JSCC) module to further compress the pruned tokens and ensure their robust transmission over error-prone channels. By employing lightweight cascaded linear layers and Signal-to-Noise Ratio (SNR)-adaptive mechanism, this module adapts to fluctuating channel conditions while maintaining minimal computational overhead. Extensive experiments on ScanQA and SQA3D datasets indicate the exceptional efficiency and robustness of our approach. Specifically, on ScanQA validation set, the LGS-HTP module could achieve 90\% pruning ratio while incurring less than a 1\% CIDEr loss. Building on this, the T-JSCC further compresses the tokens by 18$\boldsymbol{\times}$, resulting in an overall 180$\boldsymbol{\times}$ data volume reduction for robust transmission, while saving 90\% in computational complexity for LVLM deployment. Our code is available at \url{https://github.com/yuanhui0325/TP-JSCC}.
\end{abstract}
\vspace{-1.2em}
\section{Introduction}
\IEEEPARstart{L}{arge} Vision-Language Models (LVLMs) have achieved unprecedented advancements, driven by an exponential increase in computational power and the surge of large-scale datasets. These models exhibit excellent capabilities in multimodal understanding and reasoning, significantly enhancing the cognitive competence of embodied agents (EAs) \cite{ref3}. Particularly in unstructured environments requiring complex reasoning, such as 3D Visual Question Answering (3D-VQA), conventional lightweight models are limited by shallow semantic representations and often fail to capture robust cross-modal correspondences. In contrast, LVLMs leverage extensive prior knowledge to achieve superior cross-modal alignment \cite{ref4}. Consequently, integrating LVLMs into EAs is an inevitable trend toward general-purpose embodied intelligence.

However, the remarkable capabilities of LVLMs come at the cost of substantial computational overhead. Most EAs, such as intelligent drones and domestic service robots, are subject to strict physical constraints, including size, thermal limits, and battery power, rendering on-device LVLM inference computationally prohibitive. Therefore, it is necessary to develop a Cloud-Edge Collaborative Paradigm (CE-CP) for EAs \cite{ref7}. In this paradigm, EAs function as edge terminals focusing on data acquisition, compression, and transmission, while resource-abundant cloud servers execute the LVLM inference.

Although CE-CP avoids the on-device computational bottlenecks, it shifts the heavy burden to the communication channels. In error-prone communication environments, such as post-disaster rescue and deep-well exploration, communication channels are typically plagued by severe channel noise and strictly limited bandwidth. Under such conditions, directly transmitting raw high-resolution visual data becomes intractable. Specifically, the massive volume of raw pixels incurs unacceptable latency, while their inherent vulnerability to noise leads to severe transmission distortion, degrading the inference performance of LVLMs. Therefore, robust Joint Source-Channel Coding (JSCC) paradigms are necessary to ensure reliable transmission.

However, before designing the coding scheme, a fundamental question must be addressed: whether to compress raw visual pixels or their higher-level semantic representations (i.e., visual tokens). Compressing visual tokens offers three distinct advantages. First, visual encoders intrinsically filter out pixel-level noise and spatial redundancy, yielding a highly compact representation that optimizes bandwidth utilization at the source. Second, despite being distilled features, tokens retain structured correlations within the semantic space, unlocking the potential for deep compression far beyond the theoretical limits of pixel-level coding. Third, as latent representations, visual tokens are highly non-invertible, which substantially mitigates the risk of privacy leakage. Therefore, from the perspectives of both communication efficiency and privacy preservation, transmitting compressed visual tokens emerges as a highly effective paradigm \cite{ref10, ref11, ref12, ref13}.

To fully exploit the transmission potential of visual tokens over bandwidth-limited and error-prone communication channels, we propose an efficient token pruning and JSCC method for LVLM-based 3D-VQA in cloud-edge collaboration, named TP-JSCC. It effectively tackles the core bottlenecks of redundancy and distortion through two collaborative modules. First, to fundamentally eliminate token-level redundancy, we propose a Local and Global Semantics-based Hybrid Token Pruning (LGS-HTP) module. By explicitly decoupling foreground from background tokens and jointly preserving local and global semantics, this module effectively mitigates the severe performance degradation of LVLM typically induced by massive pruning. Second, to further mitigate feature-level redundancy within the pruned tokens and enhance transmission robustness, we propose a Token-based Joint Source-Channel Coding (T-JSCC) module. Constrained by the limited computational capacity of EAs, T-JSCC adopts an all-linear cascaded architecture to ensure minimal on-device overhead. Furthermore, it incorporates a Signal-to-Noise Ratio (SNR) adaptive mechanism to dynamically mitigate the impact of fluctuating channel noise, ensuring robust token reconstruction at the cloud server. The main contributions of this paper are summarized as follows.

\textbf{1) TP-JSCC method:} To address LVLM inference demands under on-device computational constraints and severely noisy, low-bandwidth communication channels, we propose an efficient token pruning and JSCC method. The edge device handles token compression, while the cloud server performs decoding and complex LVLM inference, effectively offloading the computational burden away from the edge.

\textbf{2) LGS-HTP module:} To tackle the issue of token redundancy, we propose the LGS-HTP module which first extracts foreground-background priors to decouple tokens, and then jointly selects local semantic tokens and global semantic tokens to perform pruning. It removes redundant tokens while preserving the reasoning-critical semantics.

\textbf{3) T-JSCC module:} To further reduce token redundancy and mitigate severe transmission distortion of pruned tokens over error-prone communication channels, we design a lightweight T-JSCC module based on an all-linear cascaded projection. Furthermore, we incorporate an SNR-adaptive mechanism to enable channel-aware transmission, ensuring high robustness against varying channel conditions.

The remainder of this paper is organized as follows. In Section~\ref{sec:related}, we introduce related work on 3D VQA, visual token pruning, and deep joint source-channel coding briefly. In Section~\ref{sec:formulation}, we systematically introduce the framework of TP-JSCC and mathematically formulate the token pruning and coding problem, based on which we present the proposed TP-JSCC in detail in Section~\ref{sec:method}. Experimental results and conclusions are given in Sections~\ref{sec:experiment} and \ref{sec:conclusion}, respectively.

\begin{figure*}[t]
    \centering
    \includegraphics[width=0.95\linewidth]{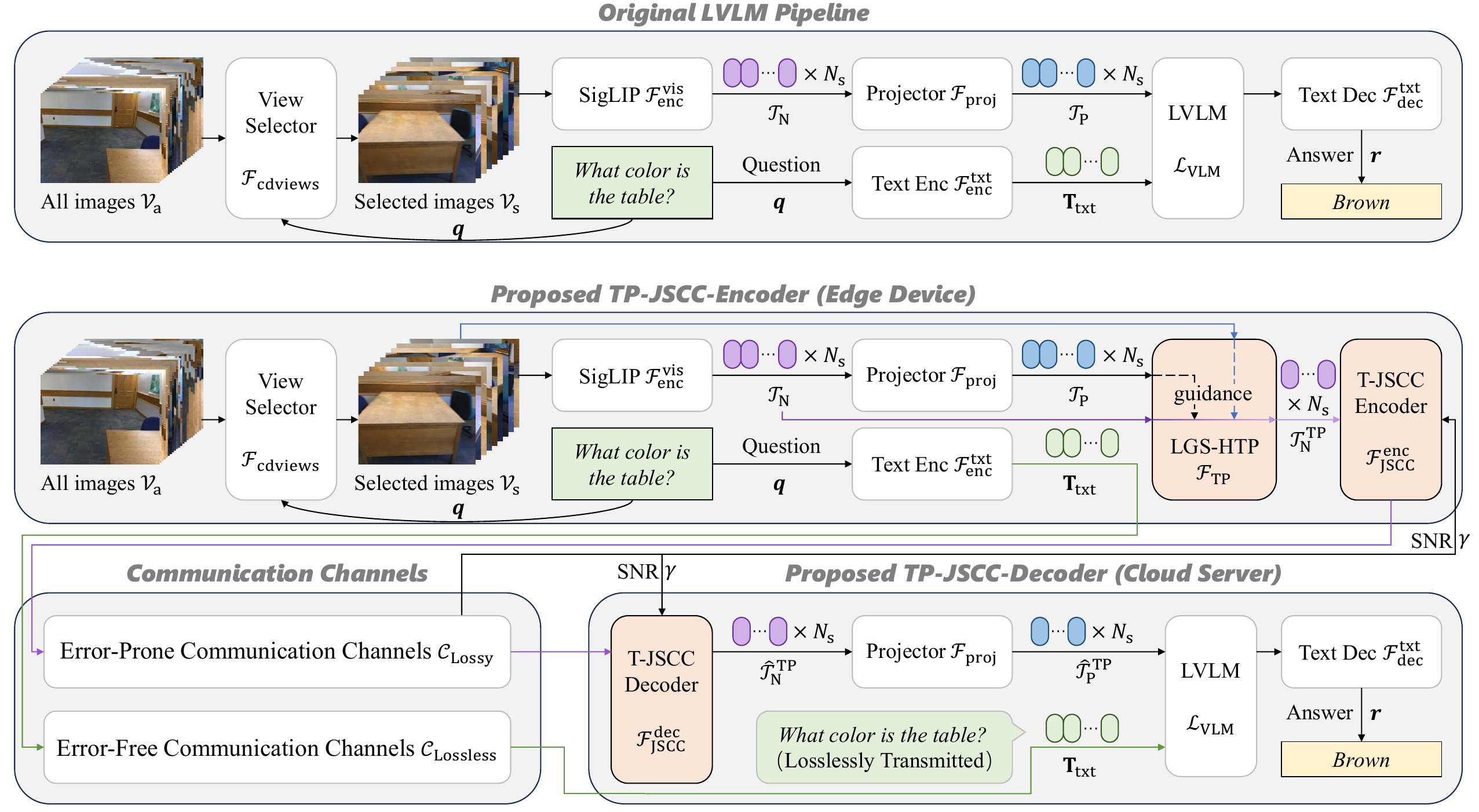}
    \vspace{-0.5em}
    \caption{Framework of the proposed Token Pruning Joint Source-Channel Coding (TP-JSCC). The LGS-HTP module effectively selects global semantic tokens to reduce visual redundancy and achieve a scalable compression ratio. Simultaneously, the T-JSCC module encodes and transmits the selected sparse tokens to maintain transmission robust to complex wireless channel conditions.}
    \label{fig1}
    \vspace{-1.2em}
\end{figure*}

\vspace{-0.5em}
\section{Related Work}
\label{sec:related}

\subsection{3D-VQA}
Existing 3D-VQA methods can be broadly categorized into three paradigms according to the representation modality of the input data: point-cloud-based methods \cite{ref14, ref15, ref17, ref18}, hybrid methods combining point clouds and multi-view images \cite{ref20, ref21, ref22, ref23}, and multi-view-image-based methods \cite{ref24, ref25, ref26, ref27}. Specifically, point-cloud and hybrid methods typically employ 3D encoders to extract high-dimensional geometric features for cross-modal fusion. However, their overall performance is severely bottlenecked by the scale and annotation quality of 3D training datasets. For instance, direct 3D point-cloud training is data-starved, as the point-cloud modality in existing high-quality datasets such as ScanQA \cite{ref14} contain only about 41,000 point-cloud-text pairs, which is orders of magnitude smaller than 2D datasets (e.g., LAION \cite{ref28} includes approximately 5.8 billion image-text pairs). Consequently, these models often struggle to establish robust cross-modal alignment, thereby constraining their generalization capabilities.

In contrast, multi-view-image-based methods directly process 2D image sequences and leverage the powerful zero-shot capabilities of pretrained LVLMs to perform 3D-VQA reasoning, effectively bypassing the point-cloud-text alignment bottleneck. A representative pipeline, LLaVA-OV \cite{ref26}, utilizes the frozen SigLIP \cite{ref29} image encoder to extract visual tokens and feeds them alongside text tokens into a pretrained Qwen \cite{ref30} model. Motivated by this architecture, cdViews \cite{ref27} introduces a lightweight attention-based view selector to replace uniform multi-view sampling, filtering for semantically relevant views to improve reasoning.

Given that frequently transmitting all multi-view visual tokens in a cloud-edge collaborative architecture inevitably introduces substantial question-irrelevant redundancy and severely degrades communication efficiency, we adopt the novel multi-view paradigm (cdViews + LLaVA-OV) as our base architecture. Utilizing view selection as a front-end filter, we construct a robust token compression framework upon this foundation to resolve the inherent transmission bottleneck.
\vspace{-1.0em}
\subsection{Visual Token Pruning}
Although LVLMs have significantly advanced 3D-VQA, their superior capability intrinsically relies on processing a massive number of visual tokens. The direct transmission of these tokens which are semantically redundant \cite{ref31}, degrades communication efficiency while simultaneously imposing an unnecessary computational burden on cloud servers \cite{ref32}. Consequently, visual token pruning has emerged as an indispensable strategy to jointly minimize both communication payload and LVLM computation overhead.

Existing token pruning approaches can be classified into text-dependent \cite{ref33, ref34, ref36, ref37} and text-independent \cite{ref38, ref39, ref40, ref41, ref42, ref43, ref44} categories. Text-dependent pruning utilizes cross-modal attention mechanisms to measure semantic alignment between visual tokens and textual queries. Although such methods theoretically preserve more task-relevant features, they face a severe operational trade-off: lightweight alignment networks often fail to establish accurate cross-modal connections, increasing the risk of irreversible removal of critical tokens \cite{ref46}; conversely, employing heavy neural networks to enhance alignment precision incurs prohibitive computational overhead \cite{ref34}, fundamentally contradicting the objective of pruning.

In contrast, text-independent methods aim to preserve maximum semantic information by assessing token importance via intrinsic statistical properties \cite{ref38}. Being predominantly training-free \cite{ref38, ref39, ref40, ref41, ref42, ref43, ref44}, these methods avoid the ambiguity of supervision labels in weakly defined pruning tasks \cite{ref37} and provide strong generalization across diverse applications. Motivated by these advantages, we adopt a text-independent strategy to optimize the balance between compression efficiency and deployment generality. 
\vspace{-1.2em}
\subsection{Deep Joint Source-Channel Coding}
Conventional communication systems founded upon Shannon's separation principle often suffer from the `cliff effect' in low-latency and finite-blocklength regimes, wherein a slight degradation in channel quality causes an abrupt performance collapse \cite{ref47, ref49, ref50, ref51}. In contrast, deep-learning-based JSCC performs end-to-end joint optimization. This paradigm demonstrates superior robustness to variations in communication channel conditions and exhibits graceful rather than catastrophic performance degradation, perfectly aligning with highly dynamic practical error-prone communication channels \cite{ref49, ref52}.

Existing deep JSCC methods are broadly categorized into human-perception oriented \cite{ref49, ref53, ref54} and machine-perception oriented \cite{ref55, ref56} approaches. The former maximizes high-fidelity pixel reconstruction (evaluated by metrics like Peak Signal-to-Noise Ratio, i.e., PSNR), which inherently preserves a massive amount of task-irrelevant information \cite{ref55}. Therefore, machine-perception JSCC aligns better with vision tasks by directly transmitting essential semantic features optimized for tasks.

Nevertheless, applying existing machine-perception JSCC to our cloud-edge LVLM framework faces two obstacles. First, there is a mismatch in representational modality: most existing methods focus on compressing CNN feature maps \cite{ref55, ref56, ref58, ref59}, and are not intrinsically designed for the discrete token sequences required by LVLMs. Second, there is a critical mismatch in computational constraints: although a few token-based JSCC schemes have recently emerged \cite{ref10, ref11, ref12, ref13}, they heavily rely on computationally intensive Transformer architectures and do not take the intra and inter token redundancy into account. Such Transformer-based designs impose unacceptable memory, communication, and computation burdens on compute-sensitive EAs in CE-CP. This precisely motivates our design of a highly streamlined, Multi-Layer Perceptron (MLP)-based Lightweight Token-JSCC method tailored for robust token transmission under strict edge constraints.

\vspace{-0.8em}
\section{Problem Formulation}
\label{sec:formulation}

The overall architecture of our proposed TP-JSCC is illustrated in Fig.~\ref{fig1}. Formally, we define the 3D-VQA task as follows. Let $\boldsymbol{q}$ denote the question and $\mathcal{S}$ denote the 3D scene. We define $\mathcal{V}_{\mathrm{a}} = \{\mathbf{I}_i\}_{i=1}^{N_{\mathrm{a}}}$ as the set of $N_{\mathrm{a}}$ view images captured from $\mathcal{S}$, where $\mathbf{I}_i$ represents the image from the $i$-th view. Given $\boldsymbol{q}$, a view selector $\mathcal{F}_{\mathrm{cdviews}}(\cdot)$ is used to choose the most informative subset of views:
\begin{equation}
\mathcal{V}_{\mathrm{s}} = \mathcal{F}_{\mathrm{cdviews}}(\mathcal{V}_{\mathrm{a}}, \boldsymbol{q}) = \{\mathbf{I}_i\}_{i=1}^{N_{\mathrm{s}}},
\end{equation}
where $N_{\mathrm{s}}$ is the number of selected views. We then employ the SigLIP \cite{ref29} image encoder $\mathcal{F}_{\mathrm{enc}}^{\mathrm{vis}}(\cdot)$ to extract multi-view native visual tokens:
\begin{equation}
\mathcal{T}_\mathrm{N} = \mathcal{F}_{\mathrm{enc}}^{\mathrm{vis}}(\mathcal{V}_{\mathrm{s}}) = \{\mathbf{T}_{\mathrm{N},i}\}_{i=1}^{N_{\mathrm{s}}},
\end{equation}
where $\mathbf{T}_{\mathrm{N},i} \in \mathbb{R}^{N_{\mathrm{p}} \times D_{\mathrm{N}}}$, $N_{\mathrm{p}}$ denotes the number of tokens per view, and $D_{\mathrm{N}}$ is the output feature dimension of $\mathcal{F}_{\mathrm{enc}}^{\mathrm{vis}}(\cdot)$. Next, a projector composed of linear layers, denoted by $\mathcal{F}_{\mathrm{proj}}(\cdot)$, is applied to obtain the projected visual tokens for the LVLM:
\begin{equation}
    \mathcal{T}_{\mathrm{P}} = \mathcal{F}_{\mathrm{proj}}(\mathcal{T}_{\mathrm{N}}) = \{\mathbf{T}_{\mathrm{P},i}\}_{i=1}^{N_{\mathrm{s}}},
\end{equation}
where $\mathbf{T}_{\mathrm{P},i} \in \mathbb{R}^{N_{\mathrm{P}} \times D_{\mathrm{P}}}$ and $D_{\mathrm{P}}$ is the hidden dimension of the LVLM. Meanwhile, the text encoder $\mathcal{F}_{\mathrm{enc}}^{\mathrm{txt}}(\cdot)$ encodes $\boldsymbol{q}$ into text tokens $\mathbf{T}_{\mathrm{txt}} \in \mathbb{R}^{N_{\mathrm{txt}} \times D_{\mathrm{P}}}$, where $N_{\mathrm{txt}}$ denotes the number of text tokens. Finally, the integrated multimodal inputs are fed into the LVLM $\mathcal{L}_{\mathrm{VLM}}(\cdot)$, followed by the text decoder $\mathcal{F}_{\mathrm{dec}}^{\mathrm{txt}}(\cdot)$ to generate the answer $\boldsymbol{r}$:
\begin{equation}
    \boldsymbol{r} = \mathcal{F}_{\mathrm{dec}}^{\mathrm{txt}}(\mathcal{L}_{\mathrm{VLM}}(\mathbf{T}_{\mathrm{txt}}, \mathcal{T}_{\mathrm{P}})).
    \label{eq:answer_generation}
\end{equation}

However, due to the extremely large parameter size of $\mathcal{L}_{\mathrm{VLM}}(\cdot)$, on-device deployment by EAs is unaffordable. This motivates us to adopt CE-CP \cite{ref7} to address the problem: the edge device (i.e., EAs) transmits $\mathbf{T}_{\mathrm{txt}}$ and $\mathcal{T}_{\mathrm{N}}$ (a transmission-friendly, more compact representation, which can be easily projected to $\mathcal{T}_{\mathrm{P}}$) to the cloud server over wireless channels, and the cloud server executes $\mathcal{L}_{\mathrm{VLM}}(\cdot)$ to generate the answer.

We assume $\mathbf{T}_{\mathrm{txt}}$ is delivered through a reliable error-free channel $\mathcal{C}_{\mathrm{Lossless}}(\cdot)$ due to its compact size and critical role in reasoning. This assumption is motivated by the widely used Unequal Error Protection (UEP) concept in communications \cite{ref60}, where key information is protected while large data allows for lossy transmission. In contrast, the much larger $\mathcal{T}_{\mathrm{N}}$ must be transmitted through a lossy channel $\mathcal{C}_{\mathrm{Lossy}}(\cdot)$. Since $\mathcal{C}_{\mathrm{Lossy}}(\cdot)$ is bandwidth-limited and noisy, direct transmission not only incurs excessive latency but also leads to severe 3D-VQA performance degradation.

To address these challenges, we propose TP-JSCC, a two-step approach consisting of pruning and JSCC. Specifically, we first utilize $\mathcal{T}_{\mathrm{P}}$ and the semantic priors in $\mathcal{V}_{\mathrm{s}}$ to guide the pruning of $\mathcal{T}_{\mathrm{N}}$, obtaining the pruned tokens $\mathcal{T}_{\mathrm{N}}^{\mathrm{TP}}$:
\begin{equation}
    \mathcal{T}_{\mathrm{N}}^{\mathrm{TP}} = \mathcal{F}_{\mathrm{TP}}(\mathcal{T}_{\mathrm{N}}, \mathcal{T}_{\mathrm{P}}, \mathcal{V}_{\mathrm{s}}, R_{\mathrm{pr}}),
\end{equation}
where $R_{\mathrm{pr}}$ denotes the pruning ratio, and $\mathcal{F}_{\mathrm{TP}}(\cdot)$ denotes the pruning algorithm. As $\mathcal{T}_{\mathrm{N}}^{\mathrm{TP}}$ is vulnerable to channel noise, we employ a JSCC encoder $\mathcal{F}_{\mathrm{JSCC}}^{\mathrm{enc}}$ and decoder $\mathcal{F}_{\mathrm{JSCC}}^{\mathrm{dec}}$, to further compress and transmit the pruned tokens $\mathcal{T}_{\mathrm{N}}^{\mathrm{TP}}$ in the edge device, reconstructing them into $\widehat{\mathcal{T}}_{\mathrm{N}}^{\mathrm{TP}}$ at the cloud server, which are then projected to $\widehat{\mathcal{T}}_{\mathrm{P}}^{\mathrm{TP}}$ (the input tokens of the $\mathcal{L}_{\mathrm{VLM}}$):
\begin{equation}
    \widehat{\mathcal{T}}_{\mathrm{N}}^{\mathrm{TP}} = \mathcal{F}_{\mathrm{JSCC}}^{\mathrm{dec}} \left( \mathcal{C}_{\mathrm{Lossy}} \left( \mathcal{F}_{\mathrm{JSCC}}^{\mathrm{enc}}( \mathcal{T}_{\mathrm{N}}^{\mathrm{TP}} | \boldsymbol{\Theta}_{\mathrm{enc}} ) \right) | \boldsymbol{\Theta}_{\mathrm{dec}} \right),
\end{equation}
\begin{equation}
    \widehat{\mathcal{T}}_{\mathrm{P}}^{\mathrm{TP}} = \mathcal{F}_{\mathrm{proj}}(\widehat{\mathcal{T}}_{\mathrm{N}}^{\mathrm{TP}}).
\end{equation}
where $\boldsymbol{\Theta}_{\mathrm{enc}}$ and $\boldsymbol{\Theta}_{\mathrm{dec}}$ denote the learnable parameters of the JSCC encoder and decoder.

The goal of pruning is to find the optimal $\mathcal{T}_{\mathrm{N}}^{\mathrm{TP}*}$ that could be projected to $\mathcal{T}_{\mathrm{P}}^{\mathrm{TP}}$ by minimizing the Kullback-Leibler (KL) divergence $\mathcal{D}_{\mathrm{KL}}$ between the output distributions of $\mathcal{L}_{\mathrm{VLM}}$ under the original input ($\mathcal{T}_{\mathrm{P}}$) and the pruned input ($\mathcal{T}_{\mathrm{P}}^{\mathrm{TP}}$):
\begin{equation}
\begin{gathered}
    \mathcal{T}_{\mathrm{N}}^{\mathrm{TP}*} = \mathop{\arg \min}\limits_{|\mathcal{T}_{\mathrm{N}}^{\mathrm{TP}}| \le K} \mathcal{D}_{\mathrm{KL}} \big( \mathcal{L}_{\mathrm{VLM}}(\mathbf{X}_1) \parallel \mathcal{L}_{\mathrm{VLM}}(\mathbf{X}_2) \big), 
\end{gathered}
\label{eq:kl_divergence}
\end{equation}
where $\mathbf{X}_1 = (\mathbf{T}_{\mathrm{txt}}, \mathcal{T}_{\mathrm{P}})$ and $\mathbf{X}_2 = (\mathbf{T}_{\mathrm{txt}}, \mathcal{F}_{\mathrm{proj}}(\mathcal{T}_{\mathrm{N}}^{\mathrm{TP}}))$, $|\cdot|$ denotes the token counting, and $K = N_{\mathrm{s}} \times N_{\mathrm{p}} \times (1 - R_{\mathrm{pr}})$ is the total token budget. Although trainable pruning methods can directly optimize the objective in Eq.~\eqref{eq:kl_divergence}, the massive parameter scale of LVLM makes gradient backpropagation exceedingly difficult and incurs prohibitively high training costs. Therefore, we opt for a training-free token pruning approach, which approximates the objective in Eq.~\eqref{eq:kl_divergence} by maximizing the semantic information retained in the pruned tokens. Specifically, cognitive science research reveals that when humans observe complex visual scenes, their perception inherently relies on integrating both local features and global context \cite{ref62}. Inspired by this cognitive mechanism, we propose that the selected tokens should capture both the local semantics (i.e., fine-grained details of objects) and the global semantics (i.e., the overall scene context). Therefore, we formulate the pruning process as finding the optimal trade-off ($N_{\mathrm{loc}}^*, N_{\mathrm{glo}}^*$) between the number of local semantic tokens $N_{\mathrm{loc}}$ and the number of global semantic tokens $N_{\mathrm{glo}}$ in $\mathcal{T}_{\mathrm{N}}^{\mathrm{TP}}$, under a fixed pruning budget $K$:
\begin{equation}
\begin{gathered}
    (N_{\mathrm{loc}}^*, \!N_{\mathrm{glo}}^*) \!=\! \arg\max \ \mathcal{U} \Big( \mathcal{F}_{\mathrm{proj}} \big( \mathcal{T}_{\mathrm{N}}^{\mathrm{TP,loc}} \!\cup\! \mathcal{T}_{\mathrm{N}}^{\mathrm{TP,glo}} \big) \Big), \\
    \text{s.t.} \ |\mathcal{T}_{\mathrm{N}}^{\mathrm{TP,loc}}| + |\mathcal{T}_{\mathrm{N}}^{\mathrm{TP,glo}}| \!=\! N_{\mathrm{loc}} \!+\! N_{\mathrm{glo}} \!=\! K
\end{gathered}
\label{eq:optimal_tradeoff}
\end{equation}
where $\mathcal{T}_{\mathrm{N}}^{\mathrm{TP,loc}}$ and $\mathcal{T}_{\mathrm{N}}^{\mathrm{TP,glo}}$ denote the pruned local and global semantic tokens. Their union yields the pruned tokens $\mathcal{T}_{\mathrm{N}}^{\mathrm{TP}}$. Moreover, $\mathcal{U}(\cdot)$ represents the task performance of $\mathcal{L}_{\mathrm{VLM}}$.

The goal of JSCC is to further compress and robustly transmit the pruned tokens over noisy wireless channels, minimizing the distortion introduced by compression and transmission on the input features $\mathcal{T}_{\mathrm{N}}^{\mathrm{TP}}$. Specifically, the objective is to ensure that the reconstructed tokens ($\widehat{\mathcal{T}}_{\mathrm{N}}^{\mathrm{TP}}$ and $\widehat{\mathcal{T}}_{\mathrm{P}}^{\mathrm{TP}}$) remain as consistent as possible with the original tokens ($\mathcal{T}_{\mathrm{N}}^{\mathrm{TP}}$ and $\mathcal{T}_{\mathrm{P}}^{\mathrm{TP}}$). To achieve this, we jointly optimize the learnable parameters of the JSCC encoder and decoder:
\begin{equation}
    \mathop{\arg \min} \limits_{\boldsymbol{\Theta}_{\mathrm{enc}}, \boldsymbol{\Theta}_{\mathrm{dec}}} \Big( \lambda_{\mathrm{N}} \mathcal{D}(\mathcal{T}_{\mathrm{N}}^{\mathrm{TP}}, \widehat{\mathcal{T}}_{\mathrm{N}}^{\mathrm{TP}}) + \lambda_{\mathrm{P}} \mathcal{D}(\mathcal{T}_{\mathrm{P}}^{\mathrm{TP}}, \widehat{\mathcal{T}}_{\mathrm{P}}^{\mathrm{TP}}) \Big),
    \label{eq:jscc_optimization}
\end{equation}
where $\mathcal{D}(\cdot, \cdot)$ denotes a distance metric (e.g., Mean Squared Error (MSE)) to measure the semantic consistency, and $\lambda_{\mathrm{N}}, \lambda_{\mathrm{P}} > 0$ are balance parameters weighting the two losses.

\vspace{-0.8em}
\section{Proposed Method}
\label{sec:method}

As illustrated in Fig.~\ref{fig1}, building upon the CE-CP defined in Section~\ref{sec:formulation}, our proposed method directly tackles the transmission bottleneck of massive visual tokens via a highly efficient two-stage strategy.

First, to fundamentally reduce the communication payload and LVLM computational overhead, we design the LGS-HTP module to achieve token-level compression. Guided by projected tokens $\mathcal{T}_{\mathrm{P}}$ and priors from a lightweight segmentation network, it performs token pruning on the native tokens $\mathcal{T}_{\mathrm{N}}$ to obtain a subset of tokens $\mathcal{T}_{\mathrm{N}}^{\mathrm{TP}}$, while preserving both local and global semantics.

Second, to further compress the tokens and transmit them over a noisy communication channel, we introduce the T-JSCC module. At the edge device, it maps the pruned tokens $\mathcal{T}_{\mathrm{N}}^{\mathrm{TP}}$ into robust communication channel symbols. At the cloud server, it reconstructs these symbols into reconstructed visual tokens $\widehat{\mathcal{T}}_{\mathrm{N}}^{\mathrm{TP}}$, which are then projected to $\widehat{\mathcal{T}}_{\mathrm{P}}^{\mathrm{TP}}$ and aligned with the reliably transmitted textual tokens $\mathbf{T}_{\mathrm{txt}}$ to execute LVLM reasoning and generate the final answer $\boldsymbol{r}$. The detailed designs of the LGS-HTP and T-JSCC modules are elaborated in Section~\ref{subsec:lgshtp} and~\ref{subsec:tjscc}, respectively.

\vspace{-1.0em}
\subsection{Local and Global Semantics-based Hybrid Token Pruning}
\label{subsec:lgshtp}

\begin{figure*}[t]
    \centering
    \includegraphics[width=0.95\linewidth]{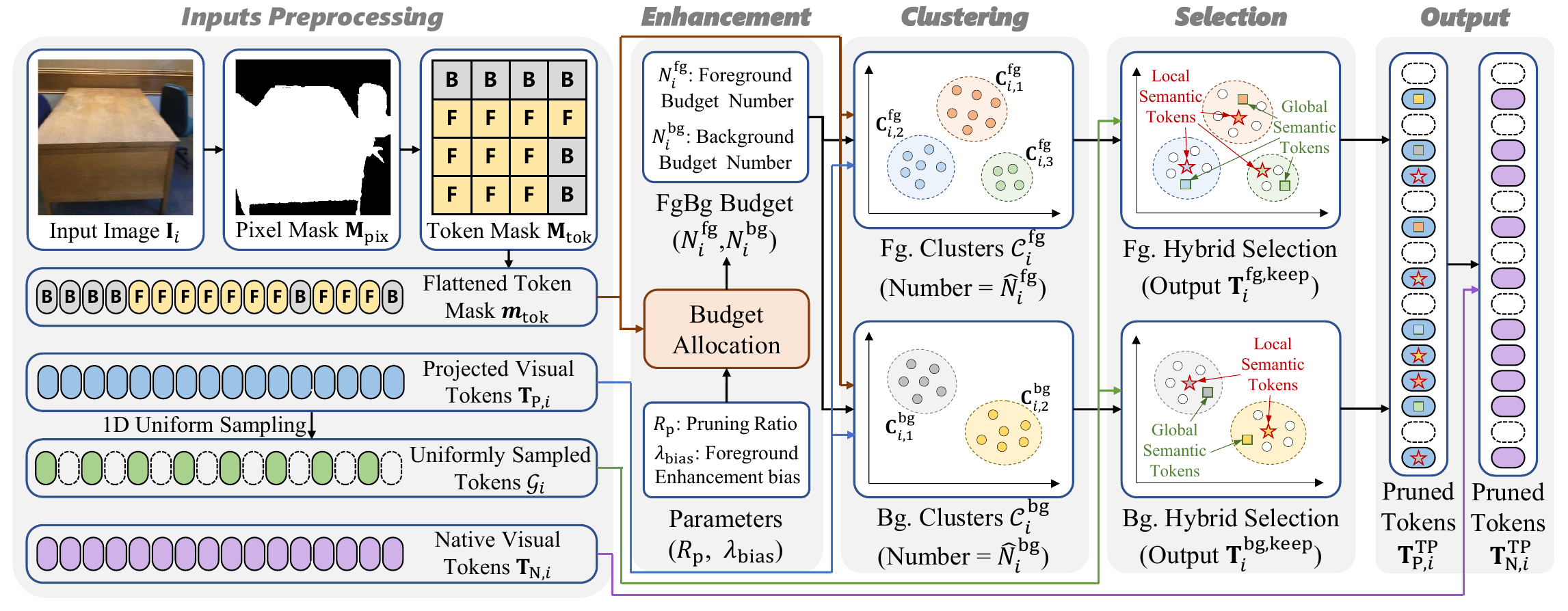}
    \vspace{-0.7em}
    \caption{Illustration of the proposed LGS-HTP module for the $i$-th view. The module comprises four steps: (1) inputs preprocessing for initial mask and token preparation, (2) enhancement for dynamic budget allocation, (3) clustering for semantic grouping, and (4) selection of the final retained tokens.}
    \label{fig2}
    \vspace{-1.3em}
\end{figure*}

The overall pipeline of the LGS-HTP module is illustrated in Fig.~\ref{fig2}. We first extract foreground-background priors (denoted as Pixel Mask in Fig.~\ref{fig2}) using a lightweight segmentation network, thereby decoupling the original tokens into foreground and background parts. Accordingly, we then introduce a foreground enhancement factor $\lambda_{\mathrm{bias}}$ to adjust the pruning budgets for foreground and background tokens, resulting in the per-view pruning budgets $(N_i^{\mathrm{fg}}, N_i^{\mathrm{bg}})$. Since the channel dimensions of input tokens vary across LVLMs with different parameter scales, performing pruning directly on the native visual tokens $\mathcal{T}_{\mathrm{N}}$ (output by the image encoder) enhances the generalization capability of the pruning model. However, the actual inputs to the LVLM are the projected tokens $\mathcal{T}_{\mathrm{P}}$. Relying exclusively on $\mathcal{T}_{\mathrm{N}}$ for pruning would inevitably neglect the inherent semantic requirements of the LVLM, thereby leading to performance degradation (as empirically verified in the ablation study in Section~\ref{ablation_2}). Therefore, we propose to first determine the pruning selection on $\mathcal{T}_{\mathrm{P}}$, and subsequently use these results to guide the pruning of $\mathcal{T}_{\mathrm{N}}$. Specifically, utilizing the per-view pruning budgets as constraints, we independently cluster the foreground and background of the projected tokens $\mathcal{T}_{\mathrm{P}}$ to construct the semantic structure. Based on this, we then jointly select the local and global semantic tokens in $\mathcal{T}_{\mathrm{P}}$, yielding $\mathcal{T}_{\mathrm{P}}^{\mathrm{TP}}$, which are then used to guide the pruning of $\mathcal{T}_{\mathrm{N}}$, yielding the final pruned token set $\mathcal{T}_{\mathrm{N}}^{\mathrm{TP}}$. The detailed designs of these steps are elaborated in Sections~\ref{lgshtp_1} and~\ref{lgshtp_2}.

\subsubsection{Foreground-Background Token Decoupling and Budget Allocation}
\label{lgshtp_1}

In multi-view 2D images, the spatial density of visual information typically exhibits obvious non-uniformity. Foreground regions concentrate intricate spatial structures and high-level semantic attributes of the target objects, thus carrying higher information density. In contrast, background regions predominantly comprise large homogeneous areas with lower information density. Applying uniform pruning overlooks this discrepancy, inevitably discarding critical foreground cues while retaining massive background redundancy.

\begin{figure}[htbp]
\centering
\includegraphics[width=\columnwidth]{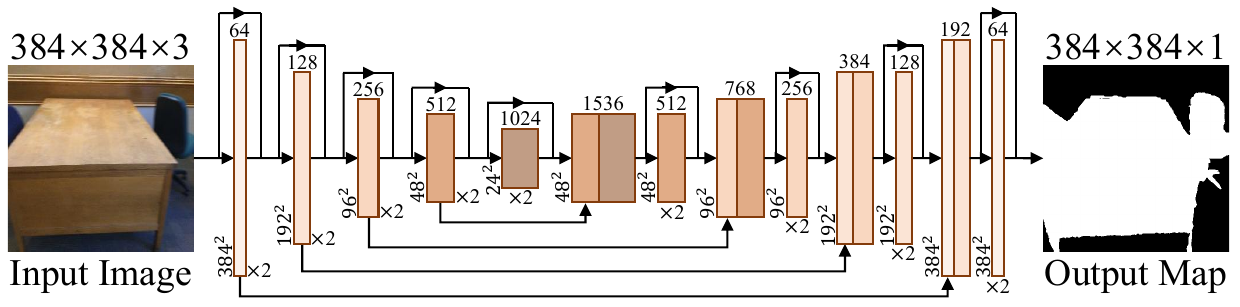}
\vspace{-2.0em}
\caption{Segmentation network for obtaining foreground-background priors.}
\vspace{-1.0em}
\label{fig3}
\end{figure}

To address this, we first employ a lightweight U-Net \cite{ref63} based segmentation network to segment the input multi-view images. As illustrated in Fig.~\ref{fig3}, this network adopts a classical encoder-decoder paradigm, where skip connections are used to fuse deep semantic features with shallow spatial details. This design efficiently produces a high-quality pixel-wise segmentation mask $\mathbf{M}_{\mathrm{pix}} \in \{0, 1\}^{384 \times 384}$ with minimal computational overhead. Here, `1' denotes the foreground region and `0' denotes the background region.

To align this pixel-level spatial prior $\mathbf{M}_{\mathrm{pix}}$ with the token space, we downsample it to the same resolution as the visual tokens and flatten it into a 1D vector, obtaining a token-level binary mask $\boldsymbol{m}_{\mathrm{tok}} \in \{0,1\}^{N_{\mathrm{p}} \times 1}$, where $N_{\mathrm{p}}$ is the original token count. Based on $\boldsymbol{m}_{\mathrm{tok}}$, we decouple the projected visual tokens ($\mathcal{T}_{\mathrm{P}}$) into foreground tokens ($\mathcal{T}_{\mathrm{P}}^{\mathrm{fg}}$) and background tokens ($\mathcal{T}_{\mathrm{P}}^{\mathrm{bg}}$), thereby providing vital semantic guidance for the subsequent pruning process.

Next, we adopt a non-uniform budget allocation strategy to maximize effective information. Given a target pruning ratio $R_{\mathrm{pr}}$, the total budget for the $i$-th view is $K_i = \lfloor N_{\mathrm{p}} \times (1 - R_{\mathrm{pr}}) \rfloor$, where $\lfloor \cdot \rfloor$ denotes the floor operation. Recognizing that foreground regions usually encapsulate high-frequency textures and complex geometric structures essential for reasoning, we introduce a foreground enhancement factor $\lambda_{\mathrm{bias}}$ to adaptively divide this budget. Taking the $i$-th view's tokens $\mathbf{T}_{\mathrm{P},i}$ as an example:
\vspace{-0.25em}
\begin{equation}
    N_i^{\mathrm{fg}} = \left\lfloor K_i \times \frac{|\mathbf{T}_{\mathrm{P},i}^{\mathrm{fg}}| \times \lambda_{\mathrm{bias}}}{|\mathbf{T}_{\mathrm{P},i}^{\mathrm{fg}}| \times \lambda_{\mathrm{bias}} + |\mathbf{T}_{\mathrm{P},i}^{\mathrm{bg}}|} \right\rfloor,
    \label{eq:fg_budget}
\end{equation}
\begin{equation}
    N_i^{\mathrm{bg}} = K_i - N_i^{\mathrm{fg}},
    \label{eq:bg_budget}
    \vspace{-0.25em}
\end{equation}
where $|\cdot|$ denotes the cardinality of the token set; $\mathbf{T}_{\mathrm{P},i}^{\mathrm{fg}}$ and $\mathbf{T}_{\mathrm{P},i}^{\mathrm{bg}}$ represent the decoupled foreground and background tokens of $\mathbf{T}_{\mathrm{P},i}$, respectively; $N_i^{\mathrm{fg}}$ and $N_i^{\mathrm{bg}}$ denote the final allocated budgets for the foreground and background tokens of the $i$-th view, respectively.

Overall, $\lambda_{\mathrm{bias}}$ adaptively skews the proportion of retained tokens toward the foreground within the total budget $K_i$, thereby guaranteeing the preservation of critical foreground semantics. The optimal value of $\lambda_{\mathrm{bias}}$ is determined empirically through a grid search, which will be detailed in Section~\ref{hyper_1}.

\begin{figure*}[t]
    \centering
    \includegraphics[width=0.9\linewidth]{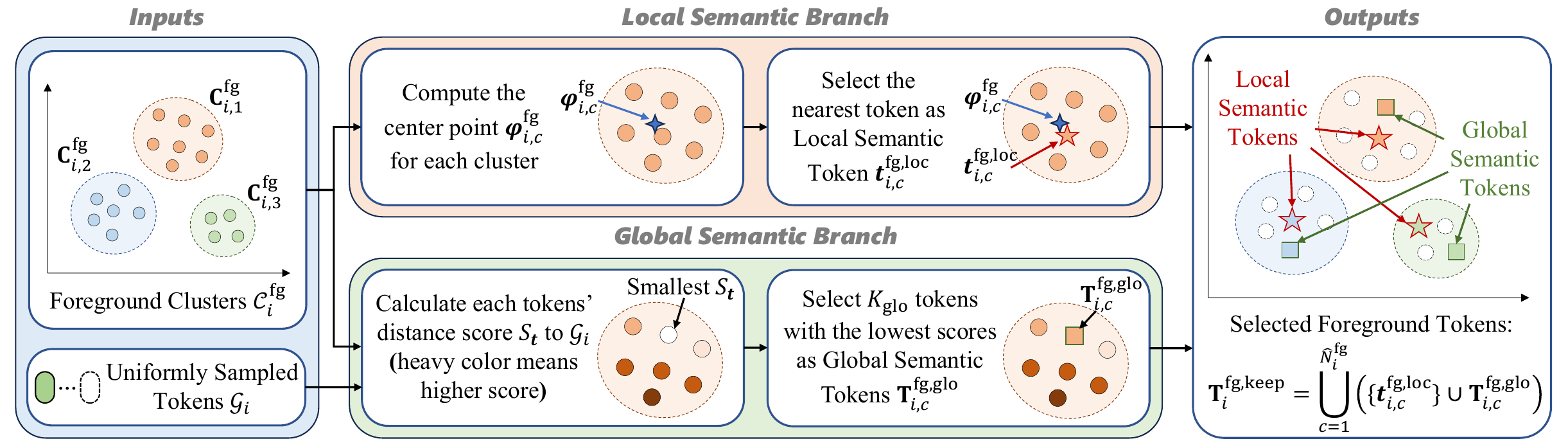}
    \vspace{-0.5em}
    \caption{Pipeline of the hybrid token selection for the foreground tokens of the $i$-th view (the background tokens are selected following the same process).}
    \label{fig4}
    \vspace{-1.0em}
\end{figure*}

\subsubsection{Local and Global Semantics-based Hybrid Token Selection}
\label{lgshtp_2}

Specifically, the proposed LGS-HTP module comprises two stages: (a) Token clustering to explicitly model the semantic hierarchies, and (b) Hybrid token selection to jointly extract local and global semantics.

\textbf{(a) Token clustering:} Semantic redundancy fundamentally lies in the strong correlations among tokens. To systematically mitigate this, we first independently cluster the foreground and background tokens (as depicted in the `Clustering' stage of Fig.~\ref{fig2}) based on their feature embeddings to characterize their underlying semantic structures, thereby providing explicit guidance for the subsequent pruning.

Specifically, taking the foreground tokens of the $i$-th view $\mathbf{T}_{\mathrm{P},i}^{\mathrm{fg}}$ as an example, we set the target number of clusters to $\widehat{N}_i^{\mathrm{fg}} = \left\lfloor \frac{N_i^{\mathrm{fg}}}{N_{\mathrm{ss}}} \right\rfloor$, where $N_{\mathrm{ss}}$ denotes the token budget allocated per cluster. We then adopt hierarchical agglomerative clustering \cite{ref65} to model the semantic cluster structure. Unlike centroid-based algorithms \cite{ref67} that rely on stochastic initialization, hierarchical clustering is deterministic and exhibits superior stability. Finally, this process partitions the foreground tokens into $\widehat{N}_i^{\mathrm{fg}}$ semantic clusters, denoted as $\mathcal{C}_i^{\mathrm{fg}} = \{\mathbf{c}_{i,c}^{\mathrm{fg}}\}_{c=1}^{\widehat{N}_i^{\mathrm{fg}}}$. The background clusters $\mathcal{C}_i^{\mathrm{bg}} = \{\mathbf{c}_{i,c}^{\mathrm{bg}}\}_{c=1}^{\widehat{N}_i^{\mathrm{bg}}}$ are obtained similarly, where $\mathbf{c}_{i,c}^{\mathrm{fg}}$ and $\mathbf{c}_{i,c}^{\mathrm{bg}}$ denote the $c$-th cluster of the $i$-th view's foreground and background tokens, respectively.

\begin{figure*}[t]
    \centering
    \includegraphics[width=0.9\linewidth]{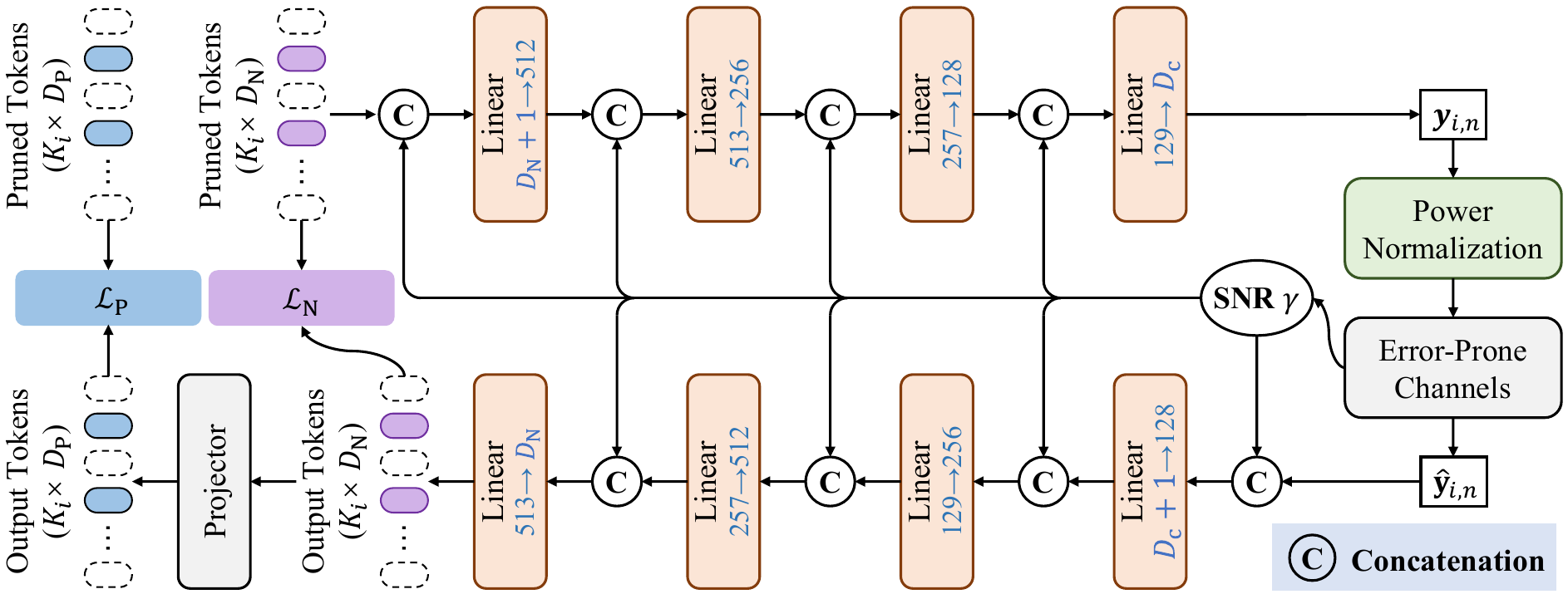}
    \vspace{-0.5em}
    \caption{Overall architecture of the proposed T-JSCC network. A lightweight MLP network first reduces the token channel dimension to 64. The compressed tokens are then power-normalized and transmitted over error-prone channels. A symmetric MLP decoder is subsequently utilized for token reconstruction. Furthermore, the channel SNR $\gamma$ is injected into both the encoding and decoding processes to achieve channel-adaptive transmission.}
    \label{fig5}
    \vspace{-1.5em}
\end{figure*}

\textbf{(b) Hybrid token selection:} As illustrated in Fig.~\ref{fig4}, given the decoupled foreground and background semantic clusters, we execute two selection branches within each cluster: a local semantic branch to extract a single local semantic token and a global semantic branch to select global semantic tokens.

In the local semantic branch, we aim to mitigate the redundancy between similar tokens. Specifically, taking $\mathbf{c}_{i,c}^{\mathrm{fg}}$ as an example, we define its local semantic centroid as the feature mean of all constituent tokens: $\boldsymbol{\varphi}_{i,c}^{\mathrm{fg}} = \frac{1}{|\mathbf{c}_{i,c}^{\mathrm{fg}}|} \sum_{\boldsymbol{t} \in \mathbf{c}_{i,c}^{\mathrm{fg}}} \boldsymbol{t}$ where $|\cdot|$ denotes the cardinality of the cluster and $\boldsymbol{t}$ represents an arbitrary token vector within $\mathbf{c}_{i,c}^{\mathrm{fg}}$. We then select the token closest to $\boldsymbol{\varphi}_{i,c}^{\mathrm{fg}}$ as the local semantic token:
\vspace{-0.5em}
\begin{equation}
    \boldsymbol{t}_{i,c}^{\mathrm{fg,loc}} = \arg \min\limits_{\boldsymbol{t} \in \mathbf{c}_{i,c}^{\mathrm{fg}}} \left\| \boldsymbol{t} - \boldsymbol{\varphi}_{i,c}^{\mathrm{fg}} \right\|_2,
    \label{eq:local_token}
    \vspace{-1.0em}
\end{equation}
where $\boldsymbol{t}_{i,c}^{\mathrm{fg,loc}}$ denotes the local semantic token of cluster $\mathbf{c}_{i,c}^{\mathrm{fg}}$. This operation precisely extracts the token that best generalize the cluster semantics, thereby removing redundancy of similar tokens. The local semantic tokens for the background clusters are obtained in the same manner.

In the global semantic branch, we aim to avoid the global semantics distortion caused by aggressive token dropping. To this end, we first select $K_i$ uniformly sampled tokens from the original tokens via 1D uniform sampling, denoted as $\mathcal{G}_i = \{\boldsymbol{t}_m^{\mathrm{glo}}\}_{m=1}^{K_i}$ where $\boldsymbol{t}_m^{\mathrm{glo}}$ denotes the $m$-th sampled token and $K_i$ is the total pruning budget for the $i$-th view. Then, conditioned on the local semantic branch, we further select $K_{\mathrm{glo}}$ global semantic tokens within each cluster whose 1D physical locations best align with the predefined uniformly sampled tokens in $\mathcal{G}_i$, where $K_{\mathrm{glo}} = N_{\mathrm{ss}} - 1$, with the $-1$ term reserving a single token budget for the local semantic token within each cluster. Notably, the optimal $K_{\mathrm{glo}}$ is experimentally tuned in Section~\ref{hyper_2}. Specifically, for $\mathbf{c}_{i,c}^{\mathrm{fg}}$, except for the local semantic token $\boldsymbol{t}_{i,c}^{\mathrm{fg,loc}}$, we define a spatial alignment score $S_{\boldsymbol{t}}$ for each remaining token $\boldsymbol{t}$ in the cluster as
\vspace{-0.2em}
\begin{equation}
    S_{\boldsymbol{t}} = \min\limits_{\boldsymbol{t}^{\mathrm{glo}} \in \mathcal{G}_i} \left| p(\boldsymbol{t}) - p(\boldsymbol{t}^{\mathrm{glo}}) \right|,
    \label{eq:spatial_score}
    \vspace{-0.5em}
\end{equation}
where $p(\cdot)$ denotes the 1D coordinate of the token. Subsequently, the $K_{\mathrm{glo}}$ tokens yielding the minimum alignment scores are selected to form the global semantic token set of the cluster:
\begin{equation}
    \mathbf{T}_{i,c}^{\mathrm{fg,glo}} = \mathrm{TopK}(\boldsymbol{s}, K_{\mathrm{glo}}),
    \label{eq:topk_global}
\end{equation}
where $\boldsymbol{s} = \{S_{\boldsymbol{t}}\}_{\boldsymbol{t} \in \mathbf{c}_{i,c}^{\mathrm{fg}}}$ denotes the set of spatial alignment scores for all remaining tokens within the cluster, and $\mathrm{TopK}(\cdot, K_{\mathrm{glo}})$ identifies the subset of tokens corresponding to the smallest alignment scores. The global semantic tokens for the background clusters are extracted analogously.

Finally, we aggregate the local and global semantic tokens for all clusters. The final pruned token sets for the foreground and background of this view are mathematically written as
\begin{equation}
    \mathbf{T}_i^{\mathrm{fg,keep}} = \bigcup_{c=1}^{\widehat{N}_i^{\mathrm{fg}}} \left( \{\boldsymbol{t}_{i,c}^{\mathrm{fg,loc}}\} \cup \mathbf{T}_{i,c}^{\mathrm{fg,glo}} \right),
    \label{eq:fg_keep}
\end{equation}
\begin{equation}
    \mathbf{T}_i^{\mathrm{bg,keep}} = \bigcup_{c=1}^{\widehat{N}_i^{\mathrm{bg}}} \left( \{\boldsymbol{t}_{i,c}^{\mathrm{bg,loc}}\} \cup \mathbf{T}_{i,c}^{\mathrm{bg,glo}} \right).
    \label{eq:bg_keep}
\end{equation}
The complete set of pruned tokens for the $i$-th view is thus $\mathbf{T}_{\mathrm{P},i}^{\mathrm{TP}} = \mathbf{T}_i^{\mathrm{fg,keep}} \cup \mathbf{T}_i^{\mathrm{bg,keep}}$. By iterating this process across all views, we obtain the multi-view pruned token set $\mathcal{T}_{\mathrm{P}}^{\mathrm{TP}}$.

Notably, while the pruning decisions are formulated in the projected token space ($D_{\mathrm{P}}$) to leverage its superior representational capacity, directly transmitting these tokens ($\mathcal{T}_{\mathrm{P}}^{\mathrm{TP}}$) incurs prohibitive communication overhead due to their significantly larger channel dimensions. Given the strict spatial correspondence between the projected and the native tokens, we directly map the corresponding indices of $\mathcal{T}_{\mathrm{P}}^{\mathrm{TP}}$ within $\mathcal{T}_{\mathrm{P}}$ onto $\mathcal{T}_{\mathrm{N}}$ to obtain pruned native tokens $\mathcal{T}_{\mathrm{N}}^{\mathrm{TP}}$, which is then fed into the subsequent T-JSCC module.

\vspace{-1.0em}
\subsection{Token Joint Source Channel Coding}
\label{subsec:tjscc}

While the LGS-HTP module effectively eliminates token-level redundancy, significant redundancy exists along the channel dimension of the pruned tokens $\mathcal{T}_{\mathrm{N}}^{\mathrm{TP}}$. To further mitigate this channel-level redundancy and ensure robust transmission under the limited computational constraints of EAs, we propose a T-JSCC module which achieves end-to-end token compression and noise robustness with strictly bounded on-device overhead.

\subsubsection{Lightweight Token-JSCC Architecture}
As illustrated in Fig.~\ref{fig5}, the proposed T-JSCC adopts a symmetric MLP-based encoder-decoder architecture. Crucially, it dynamically incorporates the channel state (SNR) $\gamma$ as a conditioning variable, which can be accurately estimated in real time using dedicated equipment and algorithms at the receiver \cite{ref69}, thereby providing channel-aware adaptivity across fluctuating channel conditions.

Specifically, for an arbitrary token in the $i$-th view, denoted as $\boldsymbol{t}_{i,n} \in \mathbb{R}^{D_{\mathrm{N}}}$, we concatenate it with $\gamma$ to form the network input $[\boldsymbol{t}_{i,n}; \gamma] \in \mathbb{R}^{D_{\mathrm{N}} + 1}$. The encoder $\mathcal{F}_{\mathrm{JSCC}}^{\mathrm{enc}}$ employs a progressive bottleneck design to reduce dimensionality, mapping the conditional input to the channel symbols:
\begin{equation}
    \boldsymbol{y}_{i,n} = \mathcal{F}_{\mathrm{JSCC}}^{\mathrm{enc}}([\boldsymbol{t}_{i,n}; \gamma]),
    \label{eq:jscc_encoder}
\end{equation}
where the dimensional transitions across the hidden layers strictly follow the configurations in Fig.~\ref{fig5}, i.e., $D_{\mathrm{N}} + 1 \rightarrow 512 \rightarrow 256 \rightarrow 128 \rightarrow D_{\mathrm{c}}$. This bottlenecked MLP minimizes computational overhead and subsequently reduces inference latency while preserving its nonlinear representation ability. Furthermore, the encoder and decoder process each token independently and in parallel with shared parameters, ensuring the overall complexity scales linearly with the number of input tokens to facilitate on-device deployment.

Subject to average power constraints \cite{ref50}, the encoded symbols are transmitted over an error-prone communication channel. The decoder $\mathcal{F}_{\mathrm{JSCC}}^{\mathrm{dec}}$ mirrors the bottleneck architecture of $\mathcal{F}_{\mathrm{JSCC}}^{\mathrm{enc}}$ and symmetrically injects the channel state $\gamma$, reconstructing each token as
\begin{equation}
    \widehat{\boldsymbol{t}}_{i,n} = \mathcal{F}_{\mathrm{JSCC}}^{\mathrm{dec}}([\widehat{\boldsymbol{y}}_{i,n}; \gamma]).
    \label{eq:jscc_decoder}
\end{equation}

After reconstructing the total native token set $\widehat{\mathcal{T}}_{\mathrm{N}}^{\mathrm{TP}}$, we employ the identical pre-trained projector $\mathcal{F}_{\mathrm{proj}}(\cdot)$ on the cloud server to linearly project it to $\widehat{\mathcal{T}}_{\mathrm{P}}^{\mathrm{TP}}$ for the subsequent LVLM reasoning.

\subsubsection{Loss Design for T-JSCC module}
Imposing a reconstruction constraint only on the native tokens ($\widehat{\mathcal{T}}_{\mathrm{N}}^{\mathrm{TP}}$) is insufficient to guarantee the reconstruction quality of the projected tokens ($\widehat{\mathcal{T}}_{\mathrm{P}}^{\mathrm{TP}}$) obtained via the projection $\mathcal{F}_{\mathrm{proj}}$. This is because neural network projections inherently amplify minor input perturbations \cite{ref70}, even slight reconstruction errors in the reconstructed native tokens can be significantly magnified by the projector.

To address this, we formulate a coarse-to-fine joint constraint. The native reconstruction loss acts as a coarse-grained constraint, stabilizing the decoded $\widehat{\mathcal{T}}_{\mathrm{N}}^{\mathrm{TP}}$ to provide a proper convergence direction. Meanwhile, the projected reconstruction loss serves as a fine-grained constraint, ensuring the reconstruction quality of the projected tokens.

Specifically, we define the coarse-grained reconstruction loss between the reconstructed native tokens $\widehat{\mathcal{T}}_{\mathrm{N}}^{\mathrm{TP}}$ and the target native tokens $\mathcal{T}_{\mathrm{N}}^{\mathrm{TP}}$ as
\begin{equation}
    \mathcal{L}_{\mathrm{N}} = \left\| \widehat{\mathcal{T}}_{\mathrm{N}}^{\mathrm{TP}} - \mathcal{T}_{\mathrm{N}}^{\mathrm{TP}} \right\|_2^2,
    \label{eq:loss_native}
\end{equation}
where $\|{\cdot}\|_2^2$ signifies the MSE. Concurrently, the fine-grained loss between the projected tokens $\widehat{\mathcal{T}}_{\mathrm{P}}^{\mathrm{TP}}$ and the targets projected tokens $\mathcal{T}_{\mathrm{P}}^{\mathrm{TP}}$ is defined as
\begin{equation}
    \mathcal{L}_{\mathrm{P}} = \left\| \widehat{\mathcal{T}}_{\mathrm{P}}^{\mathrm{TP}} - \mathcal{T}_{\mathrm{P}}^{\mathrm{TP}} \right\|_2^2.
    \label{eq:loss_projected}
\end{equation}

Finally, the overall training objective is formulated as
\begin{equation}
    \mathcal{L}_{\mathrm{JSCC}} = \lambda_{\mathrm{N}} \mathcal{L}_{\mathrm{N}} + \lambda_{\mathrm{P}} \mathcal{L}_{\mathrm{P}},
    \label{eq:loss_overall}
\end{equation}
where $\lambda_{\mathrm{N}}$ and $\lambda_{\mathrm{P}}$ are the balancing coefficients for the coarse- and fine-grained losses, respectively.

\vspace{-0.5em}
\section{Experimental Results}
\label{sec:experiment}

\subsection{Experimental Setup}
We evaluated our method on the ScanQA \cite{ref14} and SQA3D \cite{ref15} benchmarks. ScanQA provides $\sim$41K QA pairs covering diverse tasks (e.g., recognition and spatial reasoning), while SQA3D offers $\sim$33K questions emphasizing situated understanding. Given its comprehensive nature, ScanQA serves as our primary testbed for all detailed analyses and ablation studies. Meanwhile, we further evaluate our model on SQA3D to validate its generalizability.

For quantitative evaluation, we employed two complementary metrics: Top-1 Exact Match (EM) and Consensus-based Image Description Evaluation (CIDEr). Top-1 EM strictly measures exact string-matching accuracy, serving as a direct indicator of absolute correctness. Meanwhile, CIDEr evaluates n-gram similarities, providing robustness against synonymous expressions to comprehensively assess semantic consistency and linguistic quality. Throughout our experiments and hyperparameter selection, we jointly considered both metrics to optimally balance strict reasoning accuracy with textual expressiveness.

For all experiments, we employed LLaVA-OV \cite{ref26} as the foundational LVLM. As a state-of-the-art (SOTA) open-source LVLM, LLaVA-OV performs excellent in high-resolution visual representation and multi-view reasoning, making it an ideal baseline architecture to validate the effectiveness of our proposed token compression framework.

\vspace{-1.0em}
\subsection{Hyper-parameter Settings}
Our proposed method involves six user-defined hyper-parameters: the foreground-enhancement bias $\lambda_{\mathrm{bias}}$, the global semantic token number per cluster $K_{\mathrm{glo}}$, the compressed channel dimension $D_c$ in T-JSCC, the loss weights of T-JSCC $\lambda_{\mathrm{N}}$ and $\lambda_{\mathrm{P}}$, and the SNR training range for T-JSCC. In the following Section~\ref{hyper_1} to ~\ref{hyper_5}, we conduct comprehensive experiments to determine their optimal configurations.

\subsubsection{Selection of the foreground-enhancement bias \texorpdfstring{$\lambda_{\mathrm{bias}}$}{lambda\_bias}}
\label{hyper_1}

This parameter dictates the token budget allocation between the foreground and background. While a higher $\lambda_{\mathrm{bias}}$ prioritizes critical foreground semantics, an excessively high value disproportionately compresses the background, sacrificing the indispensable global context required for spatial-relation reasoning. To optimally balance foreground salience with contextual preservation, we conduct a grid search for $\lambda_{\mathrm{bias}} \in \{1.0, 1.1, \dots, 2.0\}$ under 90\% pruning (Table~\ref{tab1}).

\begin{table}[htbp]
    \centering
    \renewcommand{\arraystretch}{1.1}
    \setlength{\tabcolsep}{9mm}
    \caption{Effect of the foreground-enhancement bias $\lambda_{\mathrm{bias}}$ under 90\% token pruning on the ScanQA validation set. The best, second-best, and third-best results are highlighted in \textbf{bold}, \underline{underline}, and {\normalfont\textit{italic}}, respectively.}
    \vspace{-0.5em}
    \label{tab1}
    \begin{tabular}{ccc}
        \toprule
        \textbf{$\lambda_{\mathrm{bias}}$} & \textbf{Top-1 EM} & \textbf{CIDEr} \\
        \midrule
        1.0 & \underline{28.5989} & 98.4296 \\
        1.1 & \underline{28.5989} & \textbf{98.6438} \\
        1.2 & 28.4278 & 98.3734 \\
        1.3 & \textit{28.5775} & \textit{98.5151} \\
        1.4 & 28.5561 & 98.2152 \\
        1.5 & \textbf{28.6203} & \underline{98.5228} \\
        1.6 & 28.4492 & 98.0354 \\
        1.7 & 28.4064 & 97.8054 \\
        1.8 & 28.3209 & 97.7389 \\
        1.9 & 28.4278 & 97.7351 \\
        2.0 & 28.2995 & 97.3203 \\
        \bottomrule
    \end{tabular}
    \vspace{-1.8em}
\end{table}

As observed, performance follows an inverted-U trajectory. Lower values ($\lambda_{\mathrm{bias}} \leq 1.2$), retain excessive background redundancy, while higher values ($\lambda_{\mathrm{bias}} \geq 1.6$) compromise global spatial context. Both extremes lead to consistent drops in both metrics. Ultimately, performance peaks simultaneously at $\lambda_{\mathrm{bias}} = 1.5$, which we establish as the default configuration for all subsequent experiments.

\subsubsection{Selection of the global semantic token number per cluster \texorpdfstring{$K_{\mathrm{glo}}$}{K\_glo}}
\label{hyper_2}

As formulated in Section~\ref{sec:formulation}, precisely calibrating the number of local and global semantic tokens is critical. Since our method selects only a single local semantic token per cluster, we tune the global semantic token number per cluster, $K_{\mathrm{glo}}$, under two distinct pruning constraints (80\% and 90\% ratios) to strike the optimal balance (Table~\ref{tab2} and~\ref{tab3}).

\begin{table}[htbp]
    \vspace{-1.0em}
    \centering
    \renewcommand{\arraystretch}{1.1}
    \setlength{\tabcolsep}{8mm}
    \caption{Effect of $K_{\mathrm{glo}}$ in LGS-HTP under 80\% token pruning on the ScanQA validation set. The best, second-best, and third-best results are highlighted in \textbf{bold}, \underline{underline}, and {\normalfont\textit{italic}}, respectively.}
    \vspace{-0.5em}
    \label{tab2}
    \begin{tabular}{ccc}
        \toprule
        $K_{\mathrm{glo}}$ & \textbf{Top-1 EM} & \textbf{CIDEr} \\
        \midrule
        w/o Pruning & 29.6898 & 99.4314 \\
        0 & \underline{29.3262} & \textit{98.1047} \\
        1 & \textbf{29.7540} & \textbf{99.5614} \\
        2 & \textit{29.3048} & \underline{99.5475} \\
        3 & 28.6845 & 97.4190 \\
        4 & 28.2353 & 96.9864 \\
        \bottomrule
    \end{tabular}
    \vspace{-0.5em}
\end{table}

\begin{table}[htbp]
  \vspace{-1.5em}
    \centering
    \renewcommand{\arraystretch}{1.1}
    \setlength{\tabcolsep}{8mm}
    \caption{Effect of $K_{\mathrm{glo}}$ in LGS-HTP under 90\% token pruning on the ScanQA validation set. The best, second-best, and third-best results are highlighted in \textbf{bold}, \underline{underline}, and {\normalfont\textit{italic}}, respectively.}
    \vspace{-0.5em}
    \label{tab3}
    \begin{tabular}{ccc}
        \toprule
        $K_{\mathrm{glo}}$ & \textbf{Top-1 EM} & \textbf{CIDEr} \\
        \midrule
        w/o Pruning & 29.6898 & 99.4314 \\
        0 & \underline{28.2567} & 95.8569 \\
        1 & \textbf{28.6203} & \textbf{98.5228} \\
        2 & 27.8717 & 96.9990 \\
        3 & \textit{28.0214} & \textit{97.5871} \\
        4 & \textit{28.0214} & \underline{97.6235} \\
        \bottomrule
    \end{tabular}
    \vspace{-1.0em}
\end{table}

As observed, retaining only local semantic tokens ($K_{\mathrm{glo}} = 0$) results in insufficient global semantic coverage, which noticeably degrades performance across both metrics and causes a substantial plunge under the extreme 90\% ratio. Conversely, allocating an excessive budget to global semantic tokens (e.g., $K_{\mathrm{glo}} = 3$ or $4$) inevitably weakens key local semantics, similarly leading to performance degradation. In contrast, a moderate allocation ($K_{\mathrm{glo}} = 1$) exhibits exceptional resilience. Remarkably, at an 80\% pruning ratio, LGS-HTP even surpasses the performance of the unpruned baseline. Considering its robust stability across varying constraints, we set the default $K_{\mathrm{glo}}$ to 1, which optimally balances local and global semantics to yield peak performance.

\subsubsection{Selection of the compressed channel dimension \texorpdfstring{$D_{\mathrm{c}}$}{D\_c} in T-JSCC}
\label{hyper_3}

The compressed channel dimension $D_c$ directly dictates the transmission payload. While expanding $D_c$ enhances reconstruction fidelity, it inevitably incurs a higher communication overhead. Conversely, over-compression severely degrades semantics. To strike a rigorous balance, we evaluate varying $D_c$ configurations under 90\% pruning (Table~\ref{tab4}).

\begin{table}[htbp]
  \vspace{-1.0em}
    \centering
    \renewcommand{\arraystretch}{1.1}
    \setlength{\tabcolsep}{9.5mm}
    \caption{Impact of the compressed channel dimension $D_c$ in T-JSCC under 90\% token pruning on the ScanQA validation set.}
    \vspace{-0.5em}
    \label{tab4}
    \begin{tabular}{ccc}
        \toprule
        $D_c$ & \textbf{Top-1 EM} & \textbf{CIDEr} \\
        \midrule
        16 & 26.9733 & 92.6928 \\
        32 & 27.5722 & 94.5314 \\
        64 & 27.9144 & 96.3107 \\
        128 & 28.1711 & 96.5524 \\
        256 & 28.2995 & 97.5974 \\
        \bottomrule
    \end{tabular}
    \vspace{-1.0em}
\end{table}

\begin{figure}[htbp] 
  \vspace{-0.5em}
    \centering
    \includegraphics[width=\columnwidth]{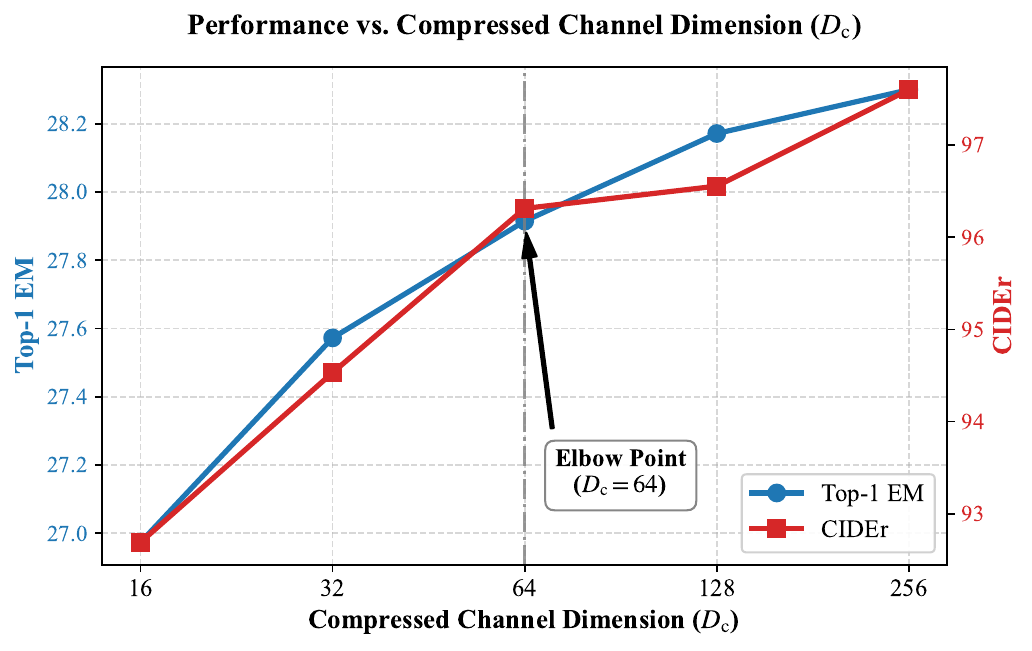} 
    \vspace{-2.0em}
    \caption{Impact of the compressed channel dimension $D_c$ on 3D-VQA performance under 90\% pruning.}
    \label{fig6}
    \vspace{-0.5em}
\end{figure}

As observed, increasing $D_c$ consistently improves both evaluation metrics, with the most significant performance gains occurring when scaling from 16 to 64. Further expanding $D_c$ to 256 yields only marginal improvements but drastically increases the data volume of the compressed tokens. Visualizing these trajectories (Fig.~\ref{fig6}) reveals a distinct inflection point (i.e., the `Elbow Point') precisely at $D_c = 64$. Consequently, to maximize reasoning accuracy while strictly control the communication payload, we establish $D_c = 64$ as the default compressed channel dimension.

\subsubsection{Selection of the loss weight of T-JSCC}
\label{hyper_4}

To determine the optimal balance between the coarse-grained reconstruction error $\mathcal{L}_{\mathrm{N}}$ and the fine-grained reconstruction error $\mathcal{L}_{\mathrm{P}}$, we evaluate weight allocations $\lambda_{\mathrm{N}}:\lambda_{\mathrm{P}}$ from 10:0 to 0:10 at $D_c = 64$ (Table~\ref{tab5}).

\begin{table}[htbp]
  \vspace{-01.5em}
    \centering
    \renewcommand{\arraystretch}{1.1}
    \setlength{\tabcolsep}{9mm}
    \caption{Effect of the T-JSCC loss-weight setting $\lambda_{\mathrm{N}}:\lambda_{\mathrm{P}}$ on the ScanQA validation set. The best, second-best, and third-best results are highlighted in \textbf{bold}, \underline{underline}, and {\normalfont\textit{italic}}, respectively.}
    \vspace{-0.5em}
    \label{tab5}
    \begin{tabular}{ccc}
        \toprule
        $\lambda_{\mathrm{N}}:\lambda_{\mathrm{P}}$ & \textbf{Top-1 EM} & \textbf{CIDEr} \\
        \midrule
        10:0 & \textbf{28.2567} & 95.0720 \\
        9:1 & 27.8289 & 95.2286 \\
        8:2 & 27.9358 & 95.4264 \\
        7:3 & 27.9144 & 95.2138 \\
        6:4 & 27.9572 & 96.0160 \\
        5:5 & 27.9144 & 96.3107 \\
        4:6 & \textbf{28.2567} & 96.6053 \\
        3:7 & 27.9786 & \underline{97.0020} \\
        2:8 & \underline{28.1497} & \textit{96.8125} \\
        1:9 & \textit{28.0428} & \textbf{97.3429} \\
        0:10 & 26.3102 & 95.0770 \\
        \bottomrule
    \end{tabular}
    \vspace{-1.8em}
\end{table}

As observed, relying exclusively on the coarse-grained constraint (10:0) maintains a reasonable Top-1 EM but severely degrades CIDEr. Conversely, entirely discarding the coarse-grained constraint (0:10) causes a precipitous drop in both metrics. These findings indicate that both fine-grained and coarse-grained losses are indispensable. Furthermore, as the weight of $\mathcal{L}_{\mathrm{P}}$ is appropriately increased, the CIDEr score exhibits a steady upward trajectory, achieving the peak performance at a ratio of 1:9. Consequently, to optimally balance training stability and semantic fidelity, we designate $\lambda_{\mathrm{N}}:\lambda_{\mathrm{P}} = 1:9$ as the default loss-weight configuration.

\subsubsection{Selection of the SNR training range for T-JSCC}
\label{hyper_5}

To determine the optimal SNR training range, we initially profile the natural 3D-VQA performance degradation over an AWGN channel without joint T-JSCC training (Table~\ref{tab6}).

\begin{table}[htbp]
  \vspace{-1.0em}
    \centering
    \renewcommand{\arraystretch}{1.1}
    \setlength{\tabcolsep}{6.5mm}
    \caption{3D-VQA performance under different AWGN channel SNRs on the ScanQA validation set.}
    \vspace{-0.5em}
    \label{tab6}
    \begin{tabular}{ccc}
        \toprule
        \textbf{SNR} & \textbf{Top-1 EM} & \textbf{CIDEr} \\
        \midrule
        w/o AWGN Channel & 28.0428 & 97.3429 \\
        0 & 7.5508 & 25.3298 \\
        5 & 12.1497 & 38.1479 \\
        10 & 18.5882 & 65.4938 \\
        15 & 23.7647 & 87.1400 \\
        20 & 26.7380 & 94.0205 \\
        25 & 27.5936 & 96.3079 \\
        30 & 27.9144 & 96.6293 \\
        \bottomrule
    \end{tabular}
    \vspace{-0.5em}
\end{table}

As observed, without channel coding, the performance suffers a catastrophic degradation under severe noise conditions (e.g., SNR = 0 dB). As the SNR increases, the performance gradually recovers and eventually saturates beyond 25 dB, indicating that the primary transmission bottleneck is concentrated within the 0$\sim$25 dB SNR regime. Consequently, we adopted 0$\sim$25 dB as the definitive SNR sampling range for training.

\vspace{-1.0em}
\subsection{Performance Comparison}

\subsubsection{LGS-HTP Performance}
We first compare the proposed LGS-HTP with three representative baseline pruning methods: Farthest Point Sampling (FPS)-based pruning, random pruning, and uniform pruning. The experimental results are summarized in Table~\ref{tab7}.

\begin{table}[htbp]
  \vspace{-1.0em}
    \centering
    \renewcommand{\arraystretch}{1.1}
    \setlength{\tabcolsep}{3mm} 
    \caption{Comparison of LGS-HTP with baseline token pruning methods at 80\% and 90\% pruning ratios on the ScanQA validation set. The best, second-best, and third-best results for each pruning ratio are highlighted in \textbf{bold}, \underline{underline}, and {\normalfont\textit{italic}}, respectively.}
    \vspace{-0.5em}
    \label{tab7}
    \begin{tabular}{cccc}
        \toprule
        \textbf{Pruning Method} & \textbf{Pruning Ratio} & \textbf{Top-1 EM} & \textbf{CIDEr} \\
        \midrule
        w/o Pruning & 0\% & 29.6898 & 99.4314 \\
        \addlinespace 
        FPS & 80\% & \underline{29.5187} & 94.0496 \\
        Random & 80\% & \textit{29.2406} & \textit{99.3210} \\
        Uniform & 80\% & 28.8770 & \textbf{101.1091} \\
        Proposed LGS-HTP & 80\% & \textbf{29.7540} & \underline{99.5614} \\
        \addlinespace 
        FPS & 90\% & \underline{28.3850} & 89.2389 \\
        Random & 90\% & \textit{28.0428} & \textit{94.7790} \\
        Uniform & 90\% & \textit{28.0428} & \underline{98.4563} \\
        Proposed LGS-HTP & 90\% & \textbf{28.6203} & \textbf{98.5228} \\
        \bottomrule
    \end{tabular}
    \vspace{-2.0em}
\end{table}

As shown, under 80\% pruning, LGS-HTP consistently achieves the best Top-1 EM while maintaining competitive CIDEr scores. Although FPS achieves relatively strong Top-1 EM by retaining richer local semantics, its lack of global semantics leads to a notable CIDEr drop. In contrast, uniform pruning yields the highest CIDEr, because it preserves more comprehensive global semantics, which provides the broad scene context essential for generating the descriptive responses that the CIDEr metric favors. However, its indiscriminate selection weakens local semantic cues, leading to the lowest Top-1 EM.

Under the more aggressive pruning rate (i.e., 90\%), the performance gap becomes more obvious, with LGS-HTP consistently exhibiting superiority over all baselines. It is worth noting that LGS-HTP overtakes uniform pruning in CIDEr scores. This is primarily because under such extreme pruning constraints, the severe loss of local semantics in uniform sampling inevitably pulls down its CIDEr metric. Conversely, our LGS-HTP inherently preserves both local and global semantics, thereby yielding superior overall results.

To comprehensively evaluate the effectiveness of our approach, we then compare LGS-HTP against several recent SOTA token pruning methods on the ScanQA validation set and the SQA3D test set. The performance under 80\% and 90\% pruning ratios is summarized in Table~\ref{tab8}.

\begin{table*}[t] 
    \centering
    \renewcommand{\arraystretch}{1.1} 
    \setlength{\tabcolsep}{8mm} 
    \caption{Comparison with SOTA pruning methods on the ScanQA validation set and the SQA3D test set. The best, second-best, and third-best results for each pruning ratio are highlighted in \textbf{bold}, \underline{underline}, and {\normalfont\textit{italic}}, respectively.}
    \vspace{-0.8em}
    \label{tab8}
    \begin{tabular}{cccccc}
        \toprule
        \multirow{2}{*}{\textbf{Pruning Method}} & \multirow{2}{*}{\textbf{Pruning Ratio}} & \multicolumn{2}{c}{\textbf{ScanQA}} & \multicolumn{2}{c}{\textbf{SQA3D}} \\
        \cmidrule(lr){3-4} \cmidrule(lr){5-6} 
        & & Top-1 EM & CIDEr & Top-1 EM & CIDEr \\
        \midrule
        w/o Pruning & 0\% & 29.6898 & 99.4314 & 50.6394 & 132.6469 \\
        \addlinespace 
        ToMe [39] & 80\% & 27.8717 & \underline{99.4191} & 48.8491 & 127.9719 \\
        GPrune [84] & 80\% & 24.6203 & 77.9843 & 46.4052 & 120.9888 \\
        CDPrune [85] & 80\% & 28.3209 & 91.8665 & 49.0480 & 128.19 \\
        DivPrune [86] & 80\% & \textit{29.1978} & \textit{93.3695} & \underline{50.0995} & \underline{131.0206} \\
        PruMerge [87] & 80\% & \underline{29.5401} & 92.1868 & \textbf{50.1563} & \textbf{131.1077} \\
        VisPrune [88] & 80\% & 28.7487 & 90.3498 & 49.6164 & 129.8927 \\
        LGS-HTP & 80\% & \textbf{29.7540} & \textbf{99.5614} & \textit{49.7869} & \textit{130.4288} \\
        \addlinespace 
        ToMe & 90\% & \textit{26.6524} & \underline{94.9143} & \textit{48.7070} & \textit{128.0065} \\
        GPrune & 90\% & 21.8182 & 67.7759 & 43.9898 & 114.3564 \\
        CDPrune & 90\% & 26.3743 & \textit{87.4624} & 48.4513 & 126.2958 \\
        DivPrune & 90\% & \underline{27.3583} & 87.3509 & \underline{49.1901} & \underline{128.5078} \\
        PruMerge & 90\% & 26.6096 & 82.4897 & 47.826 & 125.2658 \\
        VisPrune & 90\% & 26.3529 & 82.2962 & 48.3376 & 126.4189 \\
        LGS-HTP & 90\% & \textbf{28.6203} & \textbf{98.5228} & \textbf{49.5311} & \textbf{130.0889} \\
        \bottomrule
    \end{tabular}
    \vspace{-1.8em}
\end{table*}

As shown, LGS-HTP consistently establishes new state-of-the-art results on the ScanQA dataset across all metrics and pruning constraints. On the SQA3D dataset, we observe an interesting phenomenon at the 80\% pruning ratio: our method ranks third, slightly trailing behind PruMerge and DivPrune. This can be attributed to the inherent characteristics of SQA3D, where the ground-truth answers are mainly short phrases. At an 80\% pruning rate, global semantics is not yet the primary performance bottleneck, as a relatively small number of tokens can already cover the semantics required for these concise answers. Consequently, other methods that concentrate their token budget heavily on local semantics gain a marginal advantage in this specific scenario.

However, as the pruning rate increases to the extreme 90\%, where only a minimal number of tokens is retained, the lack of global semantics becomes the main bottleneck. Under this strict condition, methods that over-emphasize local semantics suffer from severe observation blind spots, leading to noticeable performance drops. In contrast, our LGS-HTP inherently preserves a balanced representation of both local and global semantics. As a result, it maintains stable performance and significantly outperforms all other SOTA methods on SQA3D at the 90\% pruning ratio, thoroughly validating the superiority and robustness of our local-global semantic theory in extreme pruning scenarios.

\subsubsection{Overall Performance}
After validating the efficacy of the LGS-HTP module, we extend our evaluation to the end-to-end performance of the integrated `LGS-HTP + T-JSCC' pipeline. Given the scarcity of token-JSCC methods specifically designed for LVLM-driven 3D-VQA, establishing direct comparisons on the ScanQA validation set is currently unfeasible. Therefore, in this section, we primarily investigate the overall effectiveness of our proposed method, as well as the impact of different SNR training strategies on the model's noise resilience.

We first evaluate a baseline strategy that uniformly samples the SNR from 0$\sim$25 dB during training, named UFM. The corresponding experimental results are presented in Table~\ref{tab9}.

\begin{table}[htbp]
    \vspace{-0.3em}
    \centering
    \renewcommand{\arraystretch}{1.1}
    \setlength{\tabcolsep}{9.5mm}
    \caption{End-to-end 3D-VQA performance of the proposed method under different SNRs with uniform SNR sampling (UFM) during training on the ScanQA validation set.}
    \vspace{-0.5em}
    \label{tab9}
    \begin{tabular}{ccc}
        \toprule
        \textbf{SNR} & \textbf{Top-1 EM} & \textbf{CIDEr} \\
        \midrule
        0 & 26.9733 & 94.0668 \\
        5 & 27.6578 & 96.0911 \\
        10 & 28.0428 & 96.4681 \\
        15 & 27.9786 & 97.1514 \\
        20 & 27.9572 & 97.0533 \\
        25 & 27.9572 & 96.5340 \\
        \bottomrule
    \end{tabular}
    \vspace{-1.0em}
\end{table}

First, a direct comparison between Table~\ref{tab6} and Table~\ref{tab9} highlights the critical necessity of our proposed T-JSCC method. Under identical SNR conditions, particularly in extreme low-SNR regimes (e.g., at 0 dB), the integration of T-JSCC yields a dramatic performance improvement over the baseline (boosting the Top-1 EM from a mere 7.55 to 26.97). This indicates that our method successfully mitigates the severe impact of channel noise on 3D-VQA performance.

Nevertheless, further analysis of Table~\ref{tab9} reveals that the UFM strategy leaves substantial room for improvement. Notably, as the SNR improves to favorable conditions (e.g., 20 and 25 dB), the performance exhibits a counter-intuitive decline rather than a continuous increase. We attribute this anomaly to the training instability caused by directly mixing high and low SNR samples. Because the reconstruction loss under extreme noise is inherently much larger than that under clean channels, jointly optimizing them with equal probability leads to severe gradient fluctuations. This prevents the model from stably converging across varying channel conditions.

To mitigate this optimization bias, we introduce a progressive (PRG) sampling strategy, adopting an anti-curriculum learning \cite{ref73} schedule. During early training epochs, we strictly confine SNR sampling to the high-noise regime (e.g., 0$\sim$5 dB), compelling the encoder to prioritize noise-robust semantic feature extraction. As training advances, we incrementally expand the sampling range epoch by epoch to encompass the full 0$\sim$25 dB spectrum. The corresponding experimental results are presented in Table~\ref{tab10}.

\begin{table}[htbp]
    \centering
    \renewcommand{\arraystretch}{1.1}
    \setlength{\tabcolsep}{9.5mm}
    \caption{End-to-end 3D-VQA performance of the proposed method under different SNRs with progressive SNR range expansion (PRG) during training on the ScanQA validation set.}
    \vspace{-0.5em}
    \label{tab10}
    \begin{tabular}{ccc}
        \toprule
        \textbf{SNR} & \textbf{Top-1 EM} & \textbf{CIDEr} \\
        \midrule
        0 & 27.2941 & 94.8344 \\
        5 & 27.8503 & 96.6230 \\
        10 & 27.8717 & 96.2943 \\
        15 & 27.8717 & 96.5267 \\
        20 & 27.9786 & 96.8218 \\
        25 & 28.1070 & 96.9992 \\
        \bottomrule
    \end{tabular}
    \vspace{-2.0em}
\end{table}

A comparative analysis between Table~\ref{tab9} and Table~\ref{tab10} reveals that the PRG strategy significantly enhances the model's low-SNR resilience. Notably, under severe channel conditions (e.g., SNR = 0 dB and 5 dB), the Top-1 EM achieves substantial improvements. Concurrently, the peak performance is firmly sustained under favorable channel conditions (e.g., SNR = 25 dB). More importantly, instead of the counter-intuitive decline at higher SNRs, the performance now exhibits a stable and consistent upward trajectory as channel conditions improve, successfully resolving the training instability observed in the UFM baseline.

\vspace{-1.2em}
\subsection{Ablation Studies}
In this section, we conduct comprehensive ablation studies to validate our proposed method. First, in Section~\ref{ablation_1}, we ablate the core components within the LGS-HTP module to verify their individual contributions. Next, to demonstrate the effectiveness of our strategy (using the projected tokens to guide the pruning of the native visual tokens), we compare the performance of different pruning schemes in Section~\ref{ablation_2}. Finally, to highlight the practical efficiency of our approach, we report the computational overhead of the LVLM before and after integrating our method in Section~\ref{ablation_3}.

\subsubsection{Ablation Studies on LGS-HTP module}
\label{ablation_1}

To rigidly validate the efficacy of the core components within the proposed LGS-HTP module, we conduct comprehensive ablation studies under the extreme 90\% pruning regime (Table~\ref{tab12}). Since local semantic extraction is indispensable for basic reasoning, we separate the ablation variables into: the foreground and background token decoupling mechanism and the initial sampling strategy for global semantics. The corresponding experimental results are presented in Table~\ref{tab11}.

\begin{table}[htbp]
  \vspace{-0.3em}
    \centering
    \renewcommand{\arraystretch}{1.1}
    \setlength{\tabcolsep}{2.5mm}
    \caption{Ablation study of LGS-HTP components: token decoupling and global semantic tokens selection strategies under 90\% token pruning on the ScanQA validation set. The best, second-best, and third-best results are highlighted in \textbf{bold}, \underline{underline}, and {\normalfont\textit{italic}}, respectively.}
    \vspace{-0.5em}
    \label{tab11}
    \begin{tabular}{cccc}
        \toprule
        \textbf{Token Decoupling} & \textbf{Selection Strategy} & \textbf{Top-1 EM} & \textbf{CIDEr} \\
        \midrule
        $\times$ & FPS & 28.5134 & 95.1825 \\
        $\times$ & Random & 28.0856 & 95.8969 \\
        $\times$ & Uniform & 28.1497 & \underline{98.0843} \\
        \checkmark & FPS & \textbf{29.1551} & 95.0144 \\
        \checkmark & Random & \textit{28.2995} & \textit{96.0001} \\
        \checkmark & Uniform & \underline{28.6203} & \textbf{98.5228} \\
        \bottomrule
    \end{tabular}
    \vspace{-2.0em}
\end{table}

As shown in Table~\ref{tab11}, explicitly decoupling tokens consistently increase performance across all selection strategies. Specifically, when paired with our uniform selection strategy, integrating the decoupling mechanism boosts Top-1 EM by 0.47 and CIDEr by 0.46 compared to the variant without decoupling (line 3). Parallel improvements are also observed under both random and FPS paradigms.

We further compare our uniform sampling against two established baselines: random sampling and FPS. Compared to random sampling, the uniform sampling yields consistent improvements, indicating that explicitly maintaining global semantics outperforms indiscriminate retention. Notably, while FPS yields a marginally higher Top-1 EM (29.16), it results in a severe drop in CIDEr (95.01) compared to the uniform strategy (98.52). We attribute this to the inherent mechanics of FPS: by greedily sampling maximally distant tokens to enforce strict diversity, it tends to over-emphasize local semantics while severely disrupting the global semantics. This omission undermines the comprehensive scene context, which is essential for CIDEr evaluation.

In contrast, the proposed uniform selection strikes the optimal balance. By strategically preserving broad global semantics without sacrificing local semantics, it ensures highly competitive Top-1 accuracy while delivering robust CIDEr performance.

\subsubsection{Effectiveness of the Guidance Strategy of LGS-HTP module}
\label{ablation_2}

To validate the effectiveness of the guidance strategy employed during the pruning process, we evaluate an unguided pruning variant that derives pruning decisions directly from the native tokens $\mathcal{T}_{\mathrm{N}}$. The corresponding performance results are reported in Table~\ref{tab12}:

\begin{table}[htbp]
  \vspace{-1.5em}
    \centering
    \renewcommand{\arraystretch}{1.1}
    \setlength{\tabcolsep}{4.5mm}
    \caption{Performance comparison between the unguided pruning variant and our proposed guidance strategy on the ScanQA validation set. The best results among the pruned variants for each pruning ratio are highlighted in \textbf{bold}.}
    \vspace{-0.5em}
    \label{tab12}
    \begin{tabular}{cccc}
        \toprule
        \textbf{Guidance} & \textbf{Pruning Ratio} & \textbf{Top-1 EM} & \textbf{CIDEr} \\
        \midrule
        $\times$ & 80\% & 29.1551 & 98.4722 \\
        \checkmark & 80\% & \textbf{29.7540} & \textbf{99.5614} \\
        \addlinespace 
        $\times$ & 90\% & 27.7219 & 97.3155 \\
        \checkmark & 90\% & \textbf{28.6203} & \textbf{98.5228} \\
        \bottomrule
    \end{tabular}
    \vspace{-1.0em}
\end{table}

As shown in Table~\ref{tab12}, the proposed guidance strategy consistently outperforms the unguided variant across different pruning ratios. These results indicate that deriving pruning decisions exclusively from the native visual tokens ($\mathcal{T}_{\mathrm{N}}$) neglects the inherent semantic requirements of the LVLM. By leveraging the projected tokens ($\mathcal{T}_{\mathrm{P}}$) for guidance, our method effectively aligns the retained visual features with the model's reasoning demands, thereby mitigating performance degradation even at high compression rates.

\subsubsection{Computational Efficiency Analysis}
\label{ablation_3}

To validate the feasibility of the proposed TP-JSCC in a cloud-edge collaborative system, this subsection reports the computational complexity of each module within TP-JSCC and the complexity of the LVLM after applying TP-JSCC. The detailed experimental results are presented in Table~\ref{tab13}:

\begin{table}[htbp]
    \vspace{-1.5em}
    \centering
    \renewcommand{\arraystretch}{1.1}
    \setlength{\tabcolsep}{1mm}
    \caption{Computational complexity (in GFLOPs) of different modules and the LVLM under various compression ratios.}
    \vspace{-0.5em}
    \label{tab13}
    \begin{tabular}{cccccc}
        \toprule
        \multirow{2}{*}{\textbf{TP-JSCC}} & \textbf{Compression} & \multirow{2}{*}{\textbf{LVLM}} & \multirow{2}{*}{\textbf{LGS-HTP}} & \multirow{2}{*}{\textbf{T-JSCC}} & \multirow{2}{*}{\textbf{Total Saving}} \\
        & \textbf{Ratio} & & & & \\
        \midrule
        $\times$ & $-$ & 93255.12 & $-$ & $-$ & $-$ \\
        \checkmark & $90\times$ & 19049.47 & 626.49 & 3.94 & 78.90\% \\
        \checkmark & $180\times$ & 9659.90 & 626.49 & 1.92 & 88.97\% \\
        \bottomrule
    \end{tabular}
    \vspace{-1.0em}
\end{table}

As shown in Table~\ref{tab13}, our proposed method effectively offloads the computationally intensive LVLM to the cloud server, incurring minimal overhead on the edge device (i.e., the combined complexity of LGS-HTP and T-JSCC is strictly bounded below 633 GFLOPs, which can be easily managed by an edge GPU like the NVIDIA Jetson TX2). Furthermore, the pruning mechanism drastically reduces the inference cost of the LVLM, which significantly lightens the computational burden on the cloud server. Notably, the 90$\times$ and 180$\times$ total compression ratios are achieved by cascading token pruning (10$\times$ for 90\% pruning, 5$\times$ for 80\%) with T-JSCC channel reduction (18$\times$, from 1152 to 64 channels), which simultaneously eases the wireless transmission pressure. Overall, our method achieves nearly 80\% and 90\% total computational savings under these two compression settings, respectively.

\vspace{-0.8em}
\section{Conclusion}
\label{sec:conclusion}

This paper focuses on a novel token-oriented CE-CP to address the severe computational constraints of EAs and the communication bottlenecks in 3D-VQA tasks over error-prone communication environments. Specifically, we proposed a highly efficient two-stage compression strategy comprising an LGS-HTP module and a T-JSCC module. By leveraging foreground-background priors and a joint selection mechanism for local and global semantic tokens, the LGS-HTP module effectively mitigates semantic redundancy. Subsequently, the T-JSCC module achieves robust, channel-aware token transmission by using an all-linear cascaded architecture and an SNR-adaptive mechanism. Extensive evaluations on the ScanQA dataset indicate the exceptional efficiency and robustness of the proposed method. Notably, our method reduces the edge-to-cloud transmission payload by approximately 180$\times$. The LGS-HTP module incurs less than a 1\% CIDEr loss at a 90\% pruning ratio, and remarkably, under extreme channel noise (SNR = 0 dB), the end-to-end CIDEr score drops by a mere 4.9\% with the help of the proposed T-JSCC module, while saving 90\% computational complexity for the LVLM. Our work provides a highly efficient and resilient token-level transmission solution, paving the way for the robust deployment of LVLM in resource-constrained EAs.

\vspace{-1.1em}
\bibliographystyle{IEEEtran}
\bibliography{TP-JSCC}

\vfill

\end{document}